\documentclass[runningheads]{llncs}

\usepackage{eccv}

\usepackage{eccvabbrv}
\usepackage{wrapfig}

\usepackage{graphicx}
\usepackage{booktabs}
\usepackage{multirow}
\usepackage[accsupp]{axessibility}  

\usepackage{hyperref}

\usepackage{orcidlink}

\begin{document}

\title{Multi-modal Relation Distillation for Unified 3D Representation Learning} 

\titlerunning{Multi-modal Relation Distillation}

\author{Huiqun Wang$^{1,2}$\thanks{For this joint project, he is also an intern with PICO ARCH; $^{\dagger}$ Corresponding author (dhuang@buaa.edu.cn)},
Yiping Bao$^{3}$,
Panwang Pan$^{3}$,
Zeming Li$^{3}$,
Xiao Liu$^{3}$ \\
Ruijie Yang$^{1,2}$ \and
Di Huang$^{1,2\dagger}$
}

\authorrunning{H. Wang et al.}

\institute{$^1$SKLSDE, Beihang University, Beijing, China\\
$^2$IRIP Lab, SCSE, Beihang University, Beijing, China\\ 
$^3$PICO}

\maketitle

\begin{abstract}
 Recent advancements in multi-modal pre-training for 3D point clouds have demonstrated promising results by aligning heterogeneous features across 3D shapes and their corresponding 2D images and language descriptions. However, current straightforward solutions often overlook intricate structural relations among samples, potentially limiting the full capabilities of multi-modal learning. To address this issue, we introduce 
\underline{\textbf{M}}ulti-modal
\underline{\textbf{R}}elation
\underline{\textbf{D}}istillation 
(\textbf{MRD}), a tri-modal pre-training framework, which is designed to effectively distill reputable large Vision-Language Models (VLM) into 3D backbones. MRD aims to capture both intra-relations within each modality as well as cross-relations between different modalities and produce more discriminative 3D shape representations. Notably, MRD achieves significant improvements in downstream zero-shot classification tasks and cross-modality retrieval tasks, delivering new state-of-the-art performance.
\end{abstract}

\section{Introduction}
Recently, 3D shape understanding has garnered increasing attention due to its wide range of applications, such as space calculation \cite{vr1,vr2,vr3,vr4}, autonomous driving \cite{ad1, ad2} and robotic perception \cite{rob1, rob2}. Despite notable advances in 3D visual analysis, the limited availability of 3D data, characterized by constraints in scale and scarcity of annotations, remains a significant barrier.

To tackle this challenge, many researchers have delved into integrating auxiliary modalities within the self-supervised learning framework. Some attempts utilize priors from the image modality to craft more instructive pretext tasks \cite{ACT, I2PMAE}, thereby enhancing the discriminative power of the learned 3D representations. Meanwhile, others focus on distilling knowledge from pre-trained models in either image or text modalities to facilitate conceptual understanding within the 3D modality \cite{PointClip, PointClip2}.

Among these endeavors, tri-modal-based methodologies \cite{CG3D, ULIP, ULIP2, OpenShape, Uni3D} have shown exceptional prowess by aligning point cloud representations with the pre-aligned image-text feature space. They employ a synthesized set of triplet data to draw point cloud representations closer to their respective image-text pairs, simultaneously distancing them from non-associated examples. The strategic use of the impressive discriminative capabilities of CLIP \cite{CLIP} markedly boosts the zero-shot performance and effectively creates a unified representation spanning different modalities, which highlights the potential of harnessing complex, multi-modal interactions to improve understanding of 3D shapes. 

\begin{figure}[!tb]
\begin{center}
\end{center}
	\includegraphics[width = 1.0\linewidth]{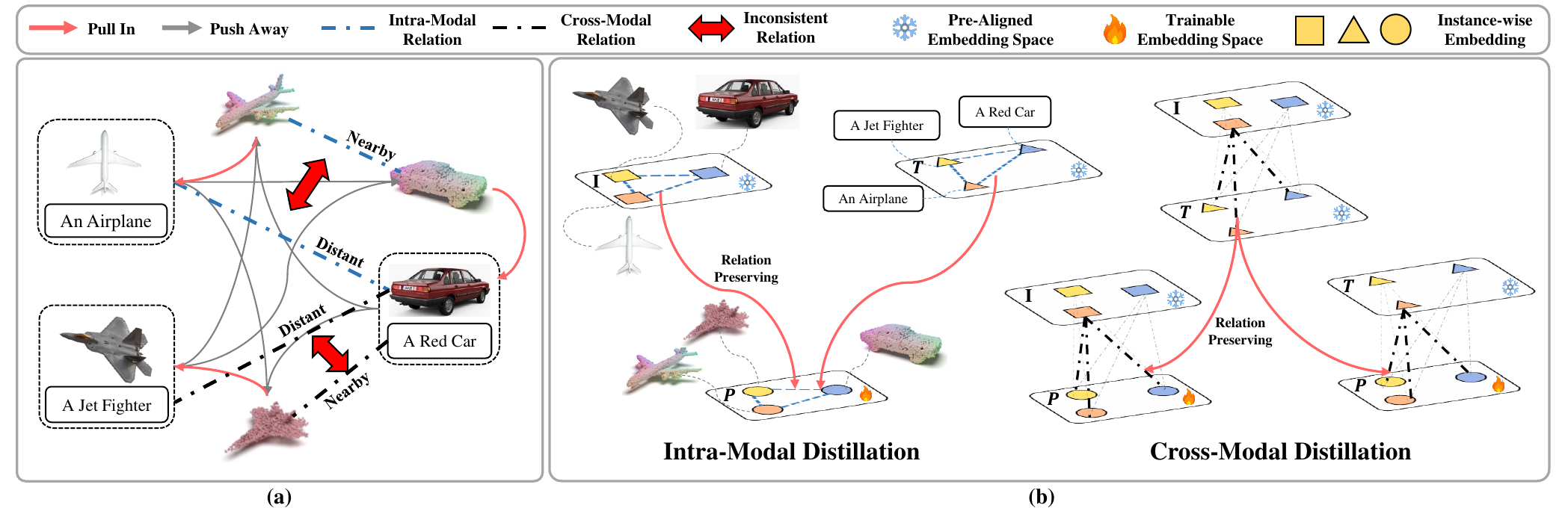}
   \caption{Illustration of Multi-modal Relation Distillation (MRD). (a) Conventional contrastive learning focuses on instance-level alignment but disrupts the intra-modality and cross-modality relations established in previous image-text alignment. For example, the nearby and distant relations between the three samples are disturbed in the 3D modality due to naive alignment. (b) MRD distills structural knowledge from both intra- and cross-modality mutual relations, aiming to preserve the semantic relations in the pre-aligned embedding spaces, thereby delivering more discriminative and coherent distributions. Zoom in for better view.}
\label{fig:fig1}
\end{figure}

As illustrated in Fig. \ref{fig:fig1}, these methodologies concentrate exclusively on aligning point cloud representations with corresponding features in image and text modalities at the individual instance level, thereby neglecting the intricate structural relations within the pre-aligned image-text feature space. Unfortunately, these mutual relations between data samples are crucial for developing a discriminative representation \cite{RKD}. For instance, intra-modal relations emphasize similarities such as shapes or textures in images, or connections among scene compositions in texts. Meanwhile, cross-modal relations reveal semantic relevance across modalities, underscoring the connections between diverse entities. The failure to consider these multifaceted relations results in the underutilization of the extensive priors learned by CLIP, culminating in a fragmented comprehension of multi-modal representations.

In this study, we aim to transfer mutual relational knowledge from the pre-aligned image-text domain into the 3D modality, thus improving the effectiveness of current multi-modal contrastive 3D pre-training approaches. Achieving this involves addressing two pivotal challenges. On the one hand, the method, which accurately models mutual relations across various modalities, has not yet been sufficiently explored. Unlike the dual-distribution scenarios of prior studies \cite{RKD, CyClip, gao2022pyramidclip}, adding the 3D modality to the pre-aligned image-text framework incurs a complex array of mutual relations that demands substantial analysis and elaborate modeling. On the other hand, inherent semantic variances within each modality create a pronounced gap between image and text modalities \cite{liang2022mind, wang2024connecting}, potentially leading to conflicts when aligning 3D representations to pre-trained image or text features. Identifying an effective strategy to mitigate such conflicts for improved convergence is an area that warrants further efforts.

To address the issues aforementioned, we commence with an in-depth analysis of existing representations and associated constraints of mutual relations, examining both within individual modalities (intra-modal) and across different modalities (cross-modal). Following this, we develop a data-driven mechanism to dynamically reconcile these conflicting intra-modal and cross-modal discrepancies. Integrating both parts, we propose a \underline{M}ulti-modal \underline{R}elation \underline{D}istillation pre-training framework, namely \textbf{MRD}, for 3D representation learning. Since MRD adeptly transfers structural relational knowledge from the image-text modality to the 3D one, it successfully enhances 3D representations by effectively merging information from diverse modalities, consequently leading to improved performance in downstream tasks.

To sum up, our contributions can be summarized as follows:
\begin{itemize}
\item We introduce MRD, an innovative self-supervised learning framework, which effectively distills both intra-modal and cross-modal relations, significantly improving the discriminative capability of 3D point cloud representations.

\item We perform comprehensive analysis and comparison of mutual relation representations within multi-modal 3D representation learning, and propose a data-driven strategy to resolve the discrepancies in the pre-aligned image-text space of CLIP.

\item We present state-of-the-art pre-trained models across diverse applications and deliver superior performance on ModelNet40 \cite{ModelNet}, ScanObjectNN \cite{ScanObj} and Objaverse \cite{Objaverse}. 
\end{itemize}

\section{Related Work}
\subsection{3D Self-supervised Learning}

Self-Supervised Learning (SSL) for point cloud understanding initiates with the 3D modality. The researchers diligently craft a range of pretext tasks such as geometric reconstruction \cite{JigSaw, GLBiRecon, OcCo, PUGANs}, mask auto-modeling \cite{PointBert, PointMAE, PointM2AE, PointGPT}, normal estimation \cite{OriPred}, and contrastive learning \cite{PointContrast, STRL, IAE} to enhance shape representations of deep learning backbones of 3D point clouds \cite{PointNet++, PointBert, pointnext, GridNet}. Contemporary endeavors seek to augment 3D SSL by importing insights from other modalities. Attempts leveraging image-guided strategies \cite{I2PMAE}, multi-modal reconstruction \cite{ACT, JointMAE, ReCon}, and weight-sharing \cite{PiMAE, Pix4Point} are integrated into conventional frameworks such as Masked Autoencoder \cite{MAE} or contrastive learning, yielding significant improvements over single-modality methods. Despite the success in enhancing the performance of downstream applications, these methods are limited by their inability to forge more comprehensive cross-modal correlations.

Leveraging the groundbreaking achievements in learning visual concepts directly from textual descriptions through contrastive learning methods, some efforts are made to capitalize on the impressive zero-shot classification capability of CLIP \cite{CLIP} to facilitate the comprehension of 3D shapes. PointCLIP \cite{PointClip} converts 3D point clouds into multi-view images for zero-shot classification by using the pre-trained visual-text encoders of CLIP and PointCLIP-2 \cite{PointClip2} advances the methodology by refining the image projection strategies and employing Large Language Models (LLMs) to optimize prompt design, reporting improved performance. However, these techniques suffer the challenges related to information loss during 3D to 2D projection and increased computational demands due to the use of images as an intermediary, which restrict their pervasion.

Simultaneously, some alternatives directly integrate multi-modal representation learning into the 3D domain. CG3D \cite{CG3D} generates ternary pairs of image-text-points from ShapeNet and aligns point cloud features with the text and image features of CLIP, aiming to distill image and text modalities into 3D representations. ULIP \cite{ULIP} enhances this approach by incorporating more intricate multi-view renderings and varied text descriptions. Subsequent developments like ULIP-2 \cite{ULIP2} and OpenShape \cite{OpenShape} expand the ternary dataset for training, leading to superior outcomes. Uni3D \cite{Uni3D} scales up the approach by adapting a Vision Transformer (ViT) pre-trained on image datasets to 3D models, reaching a milestone of 1 billion parameters. While these advances markedly improve the zero-shot classification performance, they primarily align features at the instance level across modalities and neglect the potential of exploring inter-sample relations.

\subsection{Relation Modeling in Contrastive Learning}

Previous research demonstrates that a thorough understanding of the complex relations among samples significantly improves the acquisition of structural intricacies within representations \cite{RKD}. Recently, this detailed relation modeling has been integrated into the realm of contrastive learning. RINCE \cite{hoffmann2022ranking} enhances contrastive learning by establishing a partial order among samples, thus contributing to more precise representations of 2D images. Similarly, Soft-InfoNCE \cite{softinfonce} refines the margins of the sample distribution based on sample similarities, leading to a notably more distinct feature space.

In the context of bi-modal contrastive learning, CyCLIP \cite{CyClip} introduces the constraints based on pairwise Euclidean distances, applicable both within single modality and across different modalities. Meanwhile, CLIP-PSD \cite{andonian2022clippsd} adopts similarity-based connections to mitigate the requirement of rigid one-to-one correspondence during training. Some alternative methodologies prioritize modeling complex relations through low-level similarities \cite{gao2022pyramidclip, gao2023softclip}, coherence of local patches \cite{yao2021filip}, and diversity augmentation \cite{DBLP:conf/cvpr/Yuan0K0WMKF21}. Similar enhancements also appear in the scenarios like downstream distillation \cite{DBLP:conf/NeurIPS/KimCHLL22} and CAD representation learning \cite{multicad}. Both the strategies outperform the standard baseline, indicating their effectiveness.

Nevertheless, relational distillation has not yet been well-explored in the field of multi-modal unified representation learning for 3D point clouds. The integration of the 3D modality with the pre-aligned image-text modality incurs two significant challenges. Firstly, the expansion in the number of modalities results in an increase of inter-sample relations that need to be articulated. The methodology for modeling these intra-modality and cross-modality sample relations as well as establishing the corresponding constraints remains unclear. Secondly, the inconsistency in feature distribution of the image-text modality \cite{liang2022mind, wang2024connecting} leads to conflicting relations among these distributions, making it a dilemma to leverage them for proficient representation learning in the 3D modality.

\section{Method}
\subsection{Preliminaries}

Previous studies align 3D shape representations with the pre-trained CLIP embedding space of images and texts by applying the contrastive learning loss. Given a set of synthesized image-text-3D triplet $X=\{(x_i^I,x_i^T,x_i^P)\}_{i=1}^N$ as training samples, dedicated encoders $E^I, E^T, E^P$ for each modality process inputs in their respective domains to produce the corresponding feature embeddings $f_i^I, f_i^T, f_i^P$. Notably, $E^I$ and $E^T$, derived from the extensive pre-training of CLIP on a large-scale image-text dataset, cover a data volume substantially greater than the number of available synthesized triplets. To circumvent potential model collapse, the weights for the image branch $E^I$ and the text branch $E^T$ are kept frozen, with only the point cloud branch $E^P$ being actively trained to align 3D representations with corresponding image-text pairs. Upon acquiring the feature embeddings for all samples in the batch, alignment between 3D-to-Image and 3D-to-Text is executed following the contrastive learning procedure in CLIP:

\begin{equation}
\left\{
\label{basic_info}
\begin{array}{lr}
\mathcal{L}_{P2T}=&-\frac{1}{2}\sum_i\log\frac{\exp(f_i^P\cdot f_i^T/\tau)}{\sum_j \exp(f_i^P\cdot f_j^T/\tau)}+ \log\frac{\exp(f_i^P\cdot f_i^T/\tau)}{\sum_j \exp(f_j^P\cdot f_i^T/\tau)} \\
\mathcal{L}_{P2I}=&-\frac{1}{2}\sum_i\log\frac{\exp(f_i^P\cdot f_i^I/\tau)}{\sum_j \exp(f_i^P\cdot f_j^I/\tau)}+\log\frac{\exp(f_i^P\cdot f_i^I/\tau)}{\sum_j \exp(f_j^P\cdot f_i^I/\tau)}
\end{array}
\right.
\end{equation}

\noindent where $i, j$ represent the indices of the samples, and $\tau$ denotes a learnable temperature parameter. The training minimizes the triplet contrastive loss for the weights $\theta^P$ in the 3D branch:

\begin{equation}
    \mathcal{L}_{Align}=\mathop{\min}_{\theta^P} \frac{1}{2}(\mathcal{L}_{P2T}+\mathcal{L}_{P2I})
\end{equation}

\subsection{Multi-modal Relation Representation (MRD)}
\label{SubSec.MRP}

To distill structural knowledge from the multi-modal unified representation space, it is essential to first investigate the form of mutual relation representations between samples within this multi-modal context.

With the image encoder $E^I$ and the text encoder $E^T$ fixed during the alignment phase, the learning of the 3D representation $\{f_i^P\}_{i=1}^N$ targets identifying the optimal distribution, based on the image features $\{f_i^I\}_{i=1}^N$ and text features $\{f_i^T\}_{i=1}^N$ extracted from the training set $X$. This process strives to ensure the maximal preservation of the inherent priori knowledge within the pre-aligned image-text embedding spaces.

Therefore, delineating two primary types of relations becomes crucial. The first type is the intra-modality mutual relation, symbolized as $\psi_\mathcal{M}(\cdot)$, concentrating on the sample distribution within a specific modality $\mathcal{M}$. This singular relation underscores the representational significance of various samples within the same modality. The second type, \ie the cross-modal mutual relation, is denoted as $\phi_{(\mathcal{M}_1,\mathcal{M}_2)}(\cdot)$, delving into the binary relation to reveal the semantic links between samples across modalities $\mathcal{M}_1$ and $\mathcal{M}_2$.

In this study, we explore three widely acknowledged methods for representing mutual relations: Euclidean Distance, Normalized Similarity, and Partial Order.

\begin{figure}[!tb]
\begin{center}

\resizebox{\linewidth}{!}{
	\includegraphics[width = 1.0\linewidth]{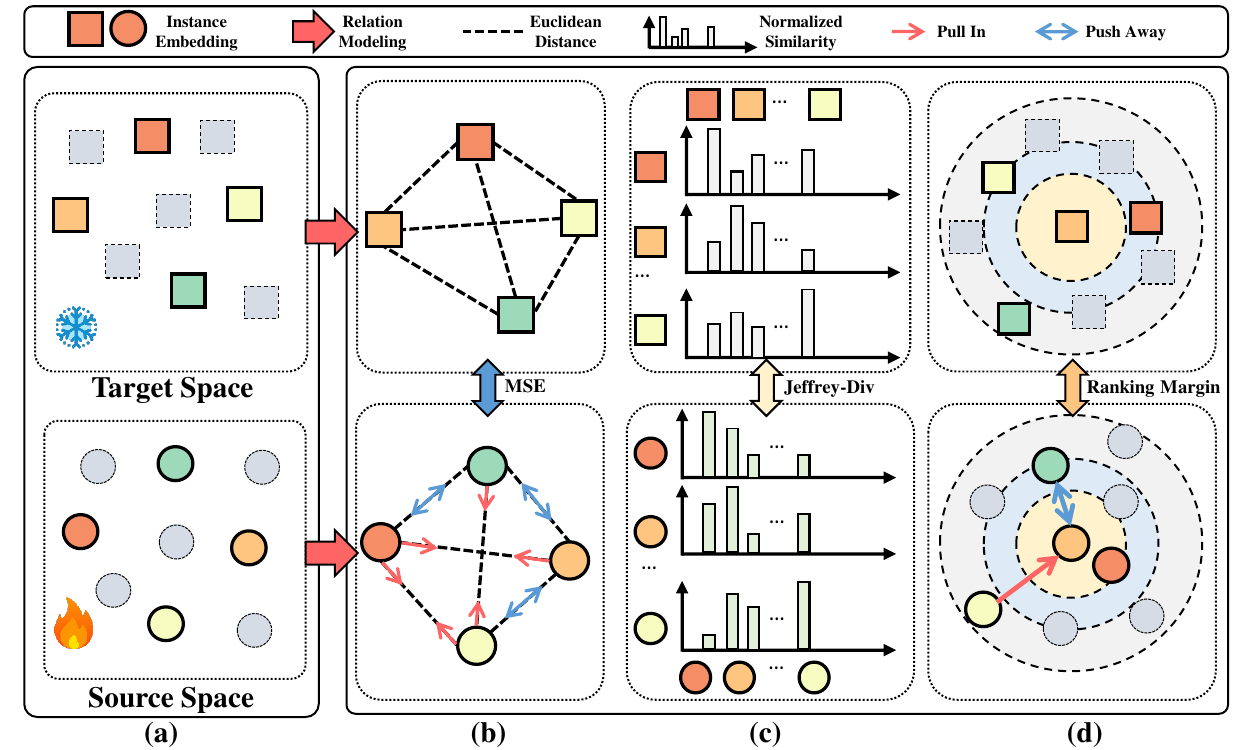}
 }
 \end{center}
   \caption{Comparison of various relation representation forms as well as corresponding distillation strategies. (a) Different embedding spaces. (b) Euclidean distance-based; (c) normalized similarity-based; and (d) partial order-based.}
\label{fig:fig2}
\end{figure}

\textbf{Euclidean Distance} serves to quantify the relative difference between two samples. For intra-modal mutual relations, this difference is captured by the pairwise distance among all samples within the same modality, denoted as:
\begin{equation}
\psi_\mathcal{M}^D(X)=\{\frac{1}{\mu}||f_i^\mathcal{M}-f_j^\mathcal{M}||^2_2\}_{(f_i^\mathcal{M},f_j^\mathcal{M})\in \mathcal{M}^2}
\end{equation}
Similarly, cross-modal relations between samples from different modalities can be expressed as follows, where $\mu$ denotes the normalization constant.
\begin{equation}
\phi_{\mathcal{M}_1,\mathcal{M}_2}^D(X)=\{\frac{1}{\mu}||f_i^{\mathcal{M}_1}-f_j^{\mathcal{M}_2}||_2^2\}_{f_i^{\mathcal{M}_1},f_j^{\mathcal{M}_2}\in \mathcal{M}_1\times\mathcal{M}_2}
\end{equation}

\textbf{Normalized Similarity} assesses correlation of features of samples within a single batch, emphasizing the overall distribution of correlation between samples rather than the absolute distance between individual sample pairs, as seen with Euclidean distance modeling. This approach offers a perspective that prioritizes relational dynamics within and across modalities. For intra-modal relations, the normalized similarity can be articulated as follows:
\begin{equation}
\psi_\mathcal{M}^S(X)=\{\frac{\exp(f_i^\mathcal{M}\cdot f_j^{\mathcal{M}}/\tau)}{\sum_{k=1}^N \exp(f_i^{\mathcal{M}}\cdot f_k^{\mathcal{M}}/\tau)}\}_{(f_i^\mathcal{M},f_j^\mathcal{M})\in \mathcal{M}^2}
\end{equation}
Similarly, the normalized similarity of inter-modal relations can be described as:
\begin{equation}
\phi_{\mathcal{M}_1,\mathcal{M}_2}^S(X)=\{\frac{\exp(f_i^{\mathcal{M}_1}\cdot f_j^{\mathcal{M}_2}/\tau)}{\sum_{k=1}^N \exp(f_i^{\mathcal{M}_1}\cdot f_k^{\mathcal{M}_2}/\tau)}\}_{f_i^{\mathcal{M}_1},f_j^{\mathcal{M}_2}\in \mathcal{M}_1\times\mathcal{M}_2}
\end{equation}

\textbf{Partial Order} captures the relative ordering between samples, differentiating itself from Euclidean distance and normalized similarity by not imposing strict metric distance constraints. Instead, it is defined through binary relations among samples. For intra-modal relations, partial order can be represented as: 
\begin{equation}
\psi_\mathcal{M}^P(X)=\{r(f_i^{\mathcal{M}},f_j^{\mathcal{M}})\}_{(f_i^\mathcal{M},f_j^\mathcal{M})\in \mathcal{M}^2}
\end{equation}
Likewise, cross-modal relations can be depicted as:
\begin{equation}
\phi_{\mathcal{M}_1,\mathcal{M}_2}^P(X)=\{r(f_i^{\mathcal{M}_1},f_j^{\mathcal{M}_2})\}_{f_i^{\mathcal{M}_1},f_j^{\mathcal{M}_2}\in \mathcal{M}_1\times\mathcal{M}_2}
\end{equation}
where $r(f_i^{\mathcal{M}_1},f_j^{\mathcal{M}_2})$ indicates the rank of $f_j^{\mathcal{M}_2}$ after sorting all samples in $\{f_k^{\mathcal{M}_2}\}_{k=1}^N$ by the Euclidean distance.

\subsection{Dynamic Relation Distillation}
\label{SubSec.DRD}

After establishing the structural relational representations, we proceed to distill intra-modality and cross-modality mutual relations. Specifically, in the context of intra-modality relations, our objective is to align the mutual relations within the 3D modality as closely as possible with those observed in the image and text modalities. Conversely, concerning cross-modal mutual relations, we aim for the 3D-to-Image and 3D-to-Text relations to emulate the correlations observed between Image-to-Text or Text-to-Image.

To tackle the issue of inconsistent mutual relations across different modalities, we introduce learnable weights for both the processes of intra-modal and cross-modal mutual relation distillation. These parameters are designed to dynamically adjust the balance between various distillation objectives in the learning phase, thus aiding in more effective network convergence. The losses associated with intra- and cross-modality distillation are articulated as follows:

\begin{equation}
\left\{
\label{mrd}
\begin{array}{llcrr}
\mathcal{L}_{Intra}^\mathcal{P} &=\alpha L(\psi_\mathcal{P}(X), \psi_\mathcal{I}(X)) + (1-\alpha) L(\psi_\mathcal{P}(X), \psi_\mathcal{T}(X)) \\
\mathcal{L}_{Cross}^{\mathcal{P}2\mathcal{T}} &=\beta L(\phi_{\mathcal{P}, \mathcal{T}}(X),\phi_{\mathcal{I}, \mathcal{T}}(X))+ (1-\beta) L(\phi_{\mathcal{P}, \mathcal{T}}(X),\phi_{\mathcal{T}, \mathcal{I}}(X))\\
\mathcal{L}_{Cross}^{\mathcal{P}2\mathcal{I}} &=\gamma L(\phi_{\mathcal{P}, \mathcal{I}}(X),\phi_{\mathcal{I}, \mathcal{T}}(X))+ (1-\gamma) L(\phi_{\mathcal{P}, \mathcal{I}}(X),\phi_{\mathcal{T}, \mathcal{I}}(X))
\end{array}
\right.
\end{equation}

Since directly incorporating parameters such as $\alpha$ and $1-\alpha$ can result in unstable optimization due to their non-smoothness, we draw inspiration from the architecture parameter settings in neural architecture search algorithms \cite{darts}. 
 
For each weight, we introduce three pairs of learnable parameters: $[w^{\alpha}_{r_1}, w^{\alpha}_{r_2};w^{\beta}_{r_1}, w^{\beta}_{r_2};\\w^{\gamma}_{r_1}, w^{\gamma}_{r_2}]$. During training, the values of $\alpha$ and $1-\alpha$ are derived from softmax of their corresponding learnable weights $[w^{\alpha}_{r_1}, w^{\alpha}_{r_2}]$. Similarly, this is applied to $\beta$ and $\gamma$, with these parameters being optimized iteratively. $L(\cdot, \cdot)$ denotes a loss function that penalizes the discrepancies between two mutual relations.

Given the distinct nature of the approaches used to characterize structural relations in Sec. \ref{SubSec.MRP}, the required form of the loss function varies accordingly. Specifically, to mod by calculating Euclidean distances, we employ the Mean Squared Error (MSE) form for distillation. The intra-modal loss can be denoted as 
\begin{equation}
L(\Gamma_1,\Gamma_2)=\frac{1}{N^2}\sum_{i,j \in (1,N)}||\gamma^{i,j}_1-\gamma^{i,j}_2||_2^2
\end{equation}
where $\Gamma_1, \Gamma_2$ represent two distinct mutual relations, and $\gamma^{i,j}_k$ denotes the Euclidean distance between the $i$-th and $j$-th samples in the relation modeling of $\Gamma_k$.

In the case of the normalized similarity form, we apply the Jeffrey divergence for imposing constraints. 
\begin{equation}
L(\Gamma_1,\Gamma_2)=\frac{1}{N}\sum_{i\in (1,N)}(KL(\gamma^{i}_1|\gamma^{i}_2)+KL(\gamma^{i}_2|\gamma^{i}_1))
\end{equation}
where $KL(\cdot|\cdot)$ represents the Kullback-Leibler Divergence, and $\gamma^{i}_k$ denotes the normalized similarity of the $i$-th sample with others in the relation modeling of $\Gamma_j$.

Lastly, for the partial order relation form, we utilize a margin-based ranking loss:
\begin{equation}
L(\Gamma_1,\Gamma_2)=\frac{1}{N^2}\sum_{i,j \in (1,N)}\max(0, -r_{i,j}*(f_i^{\mathcal{M}_1}-f_j^{\mathcal{M}_2})+\eta)
\end{equation}
where $r_{i,j}$ is the sign function indicating whether the rank order between $i$ and $j$ is consistent in $\Gamma_1$ and $\Gamma_2$, and $\eta$ represents the margin constraint.

We summarize the three relational representations and their respective constraints in Fig. \ref{fig:fig2}, and assess their impacts in the subsequent experimental section. Finally, we select the normalized similarity-based relational representation and Jeffrey divergence to distill both intra- and inter-modal relations.

\begin{figure}[!t]
\begin{center}
	\includegraphics[width = 1.0\linewidth]{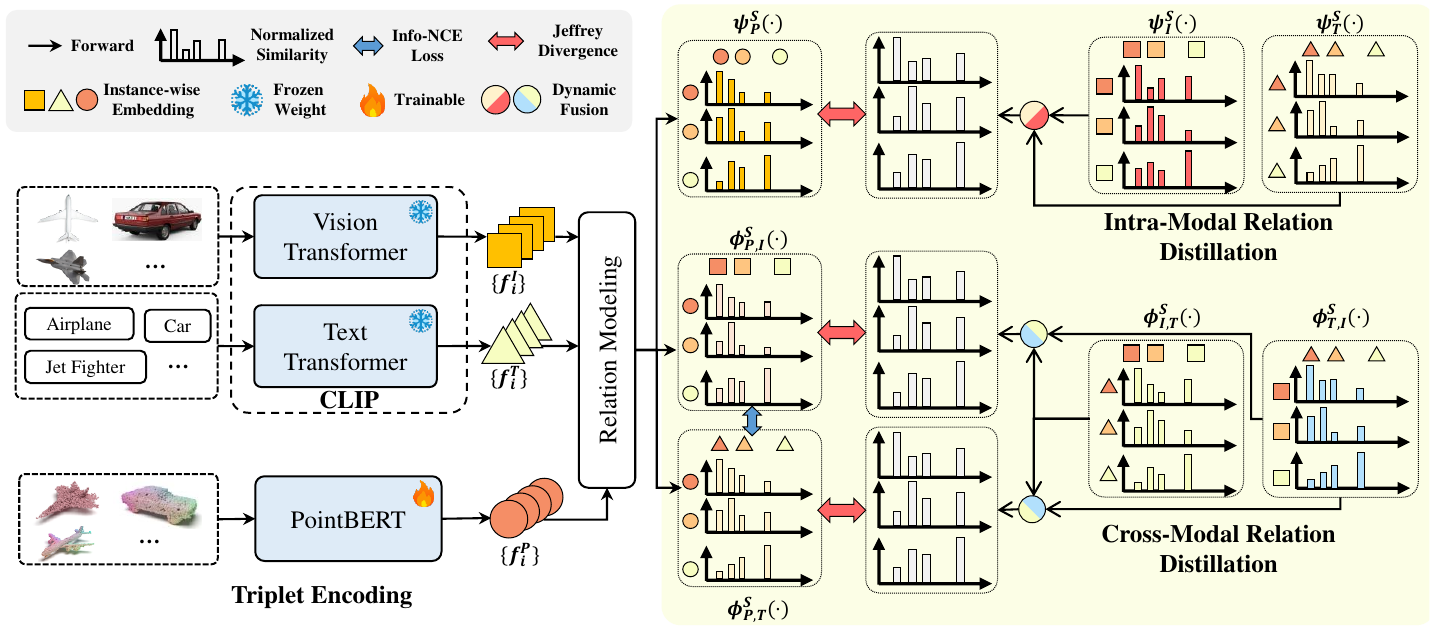}
 \end{center}
   \caption{Overall framework of MRD. With the triplet input, image-text pairs are processed by the pre-trained CLIP model, while the accompanying point clouds are encoded by the 3D encoder. MRD captures the intra-modal mutual relations $\psi(\cdot)$ within each modality and the cross-modal mutual relations $\phi(\cdot)$ across each modality pair. It then dynamically distills and transfers structural information from the pre-aligned image-text space of CLIP into the 3D representations.}
\label{fig:framework}
\end{figure}

\subsection{Framework}
The entire framework of MRD is depicted in Fig. \ref{fig:framework}. Its pre-training process involves two main components. Initially, it encodes triplet Image-Text-3D data using the pre-trained CLIP model in conjunction with a trainable 3D foundation model. Following this, both intra- and cross-modal relations between samples within each batch across the three modalities are calculated as described in Sec. \ref{SubSec.MRP}. These structural mutual relations from the image and text modalities are dynamically distilled into the 3D modality as explained in Sec. \ref{SubSec.DRD}, thereby facilitating representation learning.

During training, our objective is to minimize both the instance-level contrastive loss and the relational distillation loss across modalities. Thus, the overall loss function can be articulated as:

\begin{equation}
 \mathcal{L}=\mathop{\min}_{\theta^P} \mathcal{L}_{Align}+\lambda (\mathcal{L}_{Intra}^\mathcal{P}+\mathcal{L}_{Cross}^{\mathcal{P}2\mathcal{T}} + \mathcal{L}_{Cross}^{\mathcal{P}2\mathcal{I}})
\end{equation}
where $\lambda$ is a tunable hyper-parameter used to balance the loss terms.

\section{Experiments}
\subsection{Setup}

\textbf{Training Settings.}  In alignment with prior research \cite{Uni3D, ULIP2, OpenShape}, we compile triplets comprising 3D point clouds, 2D images, and textual descriptions from two extensive datasets of 3D objects. The first integrates the comprehensive 3D collection featured in both OpenShape and Uni3D, encompassing the datasets from Objaverse \cite{Objaverse}, ShapeNet \cite{shapenet}, 3D-FUTURE \cite{future3D}, and ABO \cite{ABO}, culminating in a total of 876K training samples. The second, ShapeNet, contains 52.5K 3D objects across 55 annotated categories. Image data for these datasets are generated through 12 preset camera angles, meticulously arranged to uniformly cover the entire spatial domain, while textual data originate from diverse sources, including curated descriptions, captions crafted by LLMs, and retrieval data. Consistent with preceding methodologies \cite{OpenShape}, we employ OpenCLIP ViT-BigG-14 as the foundational pre-trained model in this work.

\textbf{Implementation Details.}  PointBERT is taken as our fundamental model to pre-train MRD, given its status as a transformer-based backbone that has demonstrated robust performance in previous studies. To assess the impact of scaling up, we incrementally increase its size from 5M (T) to 22M (S), 32M(M), 88M (B) and 307M (L) parameters and evaluate its performance under the same settings, following the way adopted in the counterparts.

We utilize a learning rate of 0.001 for all versions of PointBERT. The batch size is configured to 192 when the model is trained on ShapeNet and increased to 512 on other datasets. Specifically, for ShapeNet, the training duration is set to 70 epochs, while for Objaverse, we extend training to 300 epochs to ensure adequate convergence. We use a cosine annealing schedule with a 15-epoch warm-up period. The weight decay is set at 0.05, $\lambda$ is set to 3, and the random seed is fixed at 0. The experiments are conducted on eight A800 GPUs, and the whole training process takes about 16 GPU-days.

\subsection{Zero-Shot Shape Classification}
For the zero-shot classification task, our evaluation considers three standard datasets: ModelNet40 \cite{ModelNet}, ScanObjectNN \cite{ScanObj}, and Objaverse \cite{Objaverse}. ModelNet40 and ScanObjectNN feature 2,468 and 2,890 test samples across 40 and 15 target classes, respectively, while Objaverse includes 46,205 test samples spanning 1,156 target classes. We calculate and compare the top-1, top-3, and top-5 accuracies with those achieved by other 3D zero-shot classification approaches. For evaluation on ModelNet and Objaverse, the input consists of sampled coordinates and color information from 10,000 points on the mesh surface of each sample. For the ScanObjectNN dataset, the input is composed of 2,048 point-cloud data points derived from the OBJ\_ONLY version.

We compare the performance of the proposed MRD framework with the representative state-of-the-art methods such as ULIP-2, OpenShape, and Uni3D. The results are presented in Tab. \ref{tab:zero}. From Tab. \ref{tab:zero} we observe that MRD achieves highly competitive results in all the settings. Notably, despite that OpenShape, Uni3D, and MRD utilize exactly the same training data, MRD significantly surpasses both OpenShape and Uni3D when operating with backbones of comparable parameter scales. It is worth noting when scaled to 88M, MRD delivers the top-1 accuracies of 53.2\% on Objaverse and 88.8\% on ModelNet40, outperforming Uni3D-L, which employs a considerably larger backbone of 307M, by 0.1\% and \textbf{2.5\%} respectively. Upon further scaling MRD-B to MRD-L, its performance improves to 53.6\%, surpassing all the competitors. Moreover, even when trained solely on ShapeNet, a much smaller dataset, MRD still consistently delivers superior results, outperforming its counterparts by a substantial margin of \textbf{1.7\%} on the most challenging Objaverse dataset.

\begin{table}[t]
  \centering
  \caption{Performance (\%) of zero-shot classification on Objaverse-LVIS~\cite{Objaverse}, ModelNet40~\cite{ModelNet}, and ScanObjectNN~\cite{ScanObj}. The best and second-best performing methods are highlighted in \textbf{bold} and \underline{underlined}, respectively.}

  \resizebox{\linewidth}{!}{
{
    \begin{tabular}{c|c|c|ccc|ccc|ccc}
    \toprule
    \multirow{2}[4]{*}{Method} &  {Params} & Training Shape  & \multicolumn{3}{c|}{Objaverse-LVIS~\cite{Objaverse}} & \multicolumn{3}{c|}{ModelNet40~\cite{ModelNet}} & \multicolumn{3}{c}{ScanObjectNN~\cite{ScanObj}} \\
\cmidrule{4-12}   &   (M)     & Source & Top1  & Top3  & Top5  & Top1  & Top3  & Top5  & Top1  & Top3  & Top5 \\
    \midrule
    PointCLIP~\cite{PointClip} & --  & \multirow{2}[1]{1.8cm}{2D Images \\at Inference} & 1.9   & 4.1      & 5.8   & 19.3  & 28.6      & 34.8  & 10.5  & 20.8      & 30.6 \\
    PointCLIP v2 \cite{PointClip2}&-- &       & 4.7   & 9.5      & 12.9  & 63.6  & 77.9      & 85.0  & 42.2  & 63.3      & 74.5 \\
    \midrule

    ReCon \cite{ReCon} &43.6 & \multirow{6}[2]{*}{ShapeNet} & 1.1      & 2.7      & 3.7      & 61.2  & 73.9      & 78.1      & 42.3      & 62.5      & 75.6 \\
    CG3D \cite{CG3D} &22.5 &       & 5.0      & 9.5       & 11.6      & 48.7  & 60.7      & 66.5      & 42.5    & 57.3      & 60.8  \\
    CLIP2Point \cite{Clip2Point} &-- &       & 2.7       & 5.8      & 7.9      & 49.5  & 71.3      & 81.2      & 25.5      & 44.6      & 59.4 \\
     ULIP+PointBERT \cite{ULIP} &22.6 &       & 6.2   & 13.6      & 17.9  & 60.4  & 79.0      & 84.4    & 51.5  & 71.1      & 80.2 \\
    OpenShape+SparseConv  \cite{OpenShape} &33.7 & & 11.6 & \underline{21.8} & \underline{27.1} & 72.9 & 87.2 & \textbf{93.0} & \underline{52.7} & 72.7 & \textbf{83.6}\\ 
     OpenShape+PointBERT &32.3  & & 10.8 & 20.2 & 25.0 & 70.3 & 86.9 & 91.3 & 51.3 & 69.4 & 78.4 \\
     \textbf{MRD-T} & 5.1  & & \underline{11.8} & 21.2 & 25.8 & \underline{74.2} & \textbf{88.3} & 90.9 & \textbf{55.7} &\textbf{ 75.5} & 83.0 \\
     \textbf{MRD-S} & 22.6 & & \textbf{13.3} & \textbf{23.8} & \textbf{29.2} & \textbf{74.2} & \underline{88.2} & \underline{92.5} & 52.1 & \underline{73.8} & \underline{81.2} \\
    \midrule

    ULIP+PointBERT  &22.6 & \multirow{2}[1]{*}{Ensembled}  & 21.4  & 38.1      & 46.0      & 71.4  & 84.4      & 89.2      & 46.0     & 66.1      & 76.4 \\
    OpenShape+SparseConv &33.7 & \multirow{2}[1]{*}{(no LVIS)} & 37.0  &   58.4    & 66.9  & 82.6  &   95.0    & 97.5  &    54.9   &  76.8     &  87.0 \\
     OpenShape+PointBERT&32.3  & & 39.1 & 60.8 & 68.9 & 85.3 & 96.2 & 97.4 & 47.2 & 72.4 & 84.7 \\
     Uni3D-B\cite{Uni3D}&88.4 & & 45.9 &  67.4   & \underline{74.8} & 86.1 & \underline{97.4}  & \underline{98.7} & 61.7& 82.0  & 89.5\\
     Uni3D-L &306.7& & \textbf{46.2} & \underline{67.6}  & 74.7 & 86.6 & 96.3 & 97.8 & 58.4& 81.4 & \underline{90.1} \\
     \textbf{MRD-S}&22.6&  & 45.2 & 67.2 & 74.7 & \textbf{88.5} & \textbf{98.2} & \textbf{99.1} & \textbf{63.0} & \textbf{84.2} & \textbf{91.9} \\
     \textbf{MRD-M}&32.3&  & \underline{45.9} & \textbf{67.8} & \textbf{75.1} & \underline{87.8} & 96.5 & 98.0 & \underline{62.3} & \underline{82.4} & 89.9\\
    \midrule
    ULIP+PointBERT &22.6 & \multirow{2}[2]{*}{Ensembled} & 26.8  & 44.8      & 52.6      & 75.1  & 88.1      & 93.2      & 51.6      & 72.5      & 82.3 \\
    OpenShape+SparseConv&33.7 &       & 43.4 & 64.8 & 72.4 & 83.4 & 95.6 & 97.8 &56.7 &   78.9    & 88.6 \\
    OpenShape+PointBERT&32.3& & 46.8 & 69.1 & 77.0 & 84.4 & 96.5 & 98.0 & 52.2 & 79.7 & 88.7 \\
    Uni3D-B \cite{Uni3D}&88.4 & & 51.7 & 74.0   & 80.8 & 86.3 & 96.5  & 97.9 & \underline{63.8} & 82.7  & 90.2 \\
    Uni3D-L&306.7 & & 53.1 & 75.0   & 81.5 & 86.3 & 96.8 & 98.3 & 58.2 & 81.8  & 89.4  \\
     \textbf{MRD-B}&88.4 & & \underline{53.2} & \underline{75.4} & \underline{82.1} & \textbf{88.8} & \textbf{97.6} & \underline{98.5} & {63.7} & \underline{84.0} & \underline{91.5} \\
      \textbf{MRD-L}&306.7& & \textbf{53.6} & \textbf{75.7} & \textbf{82.4} & \underline{87.3} & \underline{97.4} & \textbf{98.8} & \textbf{64.1} & \textbf{85.7} & \textbf{91.9} \\
    \toprule
    \end{tabular}}}
  \label{tab:zero}%
\end{table}%

\subsection{Cross-Modality Retrieval}
Considering that previous research assess cross-modal retrieval capabilities primarily relying on qualitative sample visualizations, lacking quantitative comparisons, we propose a new quantitative evaluation protocol to fill this gap for validating the effectiveness in text-to-3D retrieval. Utilizing the recently proposed Cap3D \cite{cap3D} method, which is capable of generating detailed textual descriptions for 3D data, we produce precise descriptions for the samples in the Objaverse test set. These descriptions then serve as the foundation for retrieving the most relevant samples within the Objaverse test set. Subsequently, we compute both the retrieval accuracy for individual instances and that for corresponding categories, providing a robust metric to assess the external-text retrieval capabilities of multi-modal 3D representation learning frameworks. We compare our MRD with recent state-of-the-art methods, including OpenShape and Uni3D, with the results presented in Tab. \ref{tab:retrival}.

\begin{table}[t]
  \centering
  \caption{Accuracy (\%) of external text retrieval from Cap3D to Objaverse.}
{
    \resizebox{0.65\linewidth}{!}{
    \begin{tabular}{c|ccc|ccc}
    \toprule
    \multirow{2}[1]{*}{Method}   & \multicolumn{3}{c|}{Instance-wise Accuracy} & \multicolumn{3}{c}{Category-wise Accuracy}  \\
\cmidrule{2-7}     & Top1  & Top3  & Top5  & Top1  & Top3  & Top5 \\
    \midrule
    OpenShape  & 15.4   &   28.9    &  36.8 & 49.6  &  71.1    & 78.8 \\
    Uni3D & 18.9   &  33.8   & 42.2  & 55.9  & 76.4     & 83.4  \\
    \textbf{MRD} & \textbf{20.5} & \textbf{36.0} & \textbf{44.5} & \textbf{57.9} & \textbf{78.2} & \textbf{84.8} \\
    \toprule
    \end{tabular}}}
  \label{tab:retrival}%
\end{table}%

\begin{table}[t]
  \centering
   \caption{Ablation results (\%) on various candidate relation representation forms.}
  \resizebox{\linewidth}{!}{
{
    \begin{tabular}{c|c|ccc|ccc|ccc}
    \toprule
        \multirow{2}[4]{*}{Method} & Relation  & \multicolumn{3}{c|}{Objaverse-LVIS~\cite{Objaverse}} & \multicolumn{3}{c|}{ModelNet40~\cite{ModelNet}} & \multicolumn{3}{c}{ScanObjectNN~\cite{ScanObj}} \\
\cmidrule{3-11}          & Source & Top1  & Top3  & Top5  & Top1  & Top3  & Top5  & Top1  & Top3  & Top5 \\
    \midrule
    Base & None & 11.0   & 20.0      & 24.6   & 72.2  & 85.5      & 88.8  & 53.4  & 73.5      & 81.9 \\
    \midrule
    Euclidean Distance  & \multirow{3}[2]{*}{Intra-Modal} & 11.3      & \textbf{20.3}      & 24.9       & 73.0      & 87.2  &90.8     & 52.1      & 72.8      & \textbf{84.3} \\
    Partial Order & &  10.0   &  17.7     &   21.7    &   70.6    & 86.8   &  90.9     & 53.1      & 70.8   & 79.0   \\
    Normalized Similarity & & \textbf{11.3} & 20.0 &\textbf{24.9} &\textbf{73.1} &\textbf{87.9} &\textbf{91.0} &\textbf{54.6} &\textbf{76.3} & 83.8 \\
    \midrule
        Euclidean Distance & \multirow{3}[2]{*}{Cross-Modal} & 11.0      & 20.2      & \textbf{25.1}      & \textbf{72.9}  & 86.8      & 90.1      & 53.8      & 73.1      & 81.6 \\
    Partial Order  & &  10.6     &   18.6    &  23.0     &  71.6    & \textbf{87.0} &   90.5   &  \textbf{55.2}     & 74.0   &  84.4    \\
    Normalized Similarity & & \textbf{11.4} & \textbf{20.3} & 24.9 & 72.2 & 86.7 & \textbf{91.7} & 52.4 & \textbf{75.9} & \textbf{86.2} \\
    \toprule
    \end{tabular}}}
  \label{tab:relation}%
\end{table}%

From Tab. \ref{tab:retrival}, we can find that MRD consistently outperforms its counterparts across both metrics, delivering the best performance. Benefiting from its structural distillation capability, MRD effectively learns the relations within the textual feature space. This, in turn, enhances its ability to accurately retrieve corresponding instances when processing external text descriptions, thereby demonstrating its effectiveness.

\subsection{Ablation Study}
\textbf{Candidate Relation Representation.} We evaluate the effects of different candidate relation representations and their associated distillation losses, as outlined in Sec. \ref{SubSec.MRP}, on various test sets, with the outcomes presented in Tab. \ref{tab:relation}. From Tab. \ref{tab:relation}, we can find that incorporating the relational distillation mechanism leads to improvements in accuracy for most settings. Notably, the most significant enhancement is achieved when modeling intra- and inter-modal relations using normalized similarities. This superiority may stem from the ability of normalized similarity-based relation representations, which strikes a more effective balance between the flexibility of the sample distribution and the consistency of relational constraints, in contrast to the stricter Euclidean distance and the more lenient partial order relation, thereby enhancing the performance.

\textbf{Relation Distillation Strategies.} Leveraging the normalized similarity and its corresponding distillation loss, we conduct ablation experiments on dynamic distillation, with the results presented in Tab. \ref{tab:ablation}. As observed from Tab. \ref{tab:ablation}, simultaneously introducing both intra-modal and cross-modal relation distillation leads to the performance that is less favorable compared to independently implementing either of the intra-modal or cross-modal constraints, due to the inconsistency arising from the complex spectrum of mutual relations. Notably, upon introducing the Dynamic Distillation (DD) mechanism, there is a significant gain in performance, demonstrating the efficacy of this strategy.

\begin{table*}[t]
  \centering
  \caption{Ablation results (\%) on various relation distillation strategies within MRD (IR: Intra-modal Relation, CR: Cross-modal Relation, DD: Dynamic Distillation).}
{
    \begin{tabular}{ccc|ccc|ccc|ccc}
    \toprule
    \multicolumn{3}{c|}{Setting}  & \multicolumn{3}{c|}{Objaverse-LVIS~\cite{Objaverse}} & \multicolumn{3}{c|}{ModelNet40~\cite{ModelNet}} & \multicolumn{3}{c}{ScanObjectNN~\cite{ScanObj}} \\
     \cmidrule{1-12} IR &CR&DD& Top1  & Top3  & Top5  & Top1  & Top3  & Top5  & Top1  & Top3  & Top5 \\
    \midrule
    & & & 11.0   & 20.0      & 24.6   & 72.2  & 85.5      & 88.8  & 53.4  & 73.5      & 81.9 \\
    $\checkmark$ & & & 11.3 & 20.0 &24.9 &73.1 &87.9 &91.0 &54.6 &\textbf{76.3} & 83.8 \\
     &$\checkmark$& & 11.4 & 20.3 & 24.9 & 72.2 & 86.7 & \textbf{91.7} & 52.4 & 75.9 & \textbf{86.2} \\
    \midrule
    $\checkmark$ & $\checkmark$& & 11.4  & 20.7  & 25.4  & 73.0  & 87.1  & 90.3  & 53.5  & 73.3  & 81.2  \\
    $\checkmark$ & $\checkmark$&$\checkmark$& \textbf{11.8} & \textbf{21.2} & \textbf{25.8} & \textbf{74.2} & \textbf{88.3} & 90.9 & \textbf{55.7} & 75.5 & 83.0 \\
    \toprule
    \end{tabular}}
  \label{tab:ablation}%
\end{table*}%

\begin{table}[t]
    \begin{minipage}[b]{0.45\textwidth}
        \centering
         \caption{MAE between the similarity matrices on ShapeNet.}
          \resizebox{0.9\linewidth}{!}{
       \begin{tabular}{ccc|c|c|c|c}
    \toprule
    IR &CR&DD & P2P/I2I & P2P/T2T & P2T/I2T & P2I/T2I \\
    \midrule
    & & & 0.36 & 0.20 & 0.21 & 0.17 \\
    $\checkmark$& & & 0.08 & 0.24 & 0.21 & 0.18 \\
     &$\checkmark$ & &0.37 & 0.20 & 0.19 & 0.16 \\
    \midrule
    $\checkmark$ & $\checkmark$& & 0.20 & 0.13 & 0.20 & 0.16  \\
    $\checkmark$ & $\checkmark$&$\checkmark$& 0.15 & 0.14 & 0.20 & 0.16 \\
    \bottomrule
    \end{tabular}}

        \label{tab:mae}
    \end{minipage}
    \hfill
    \begin{minipage}[b]{0.5\textwidth}
        \centering
        \caption{Performance (\%) when scaling up the model size in MRD.}
          \resizebox{\linewidth}{!}{
         \begin{tabular}{c|ccc|c|cc}
    \toprule
    Model & Depth   &   Width   &  Heads  &  Params  &   MNet40    & O-LVIS    \\
    \midrule
    MRD-T & 6   & 256     &  4  & 5.1M  &   86.7    &  47.6    \\
    MRD-S & 12   & 394     &  6  & 22.6M  &  88.6     & 51.4  \\
    MRD-M &  12  & 512     &   8 & 32.3M  &   88.0    & 52.5  \\
    MRD-B &  12  & 768     &   12 & 88.4M  &  \textbf{88.8}     & 53.2   \\
    MRD-L &  24  &  1024    &    16 & 306.7M  &  87.3     &  \textbf{53.6}  \\
    \toprule
    \end{tabular}}
        
        \label{tab:scaleup}
    \end{minipage}
\end{table}

To quantify the improvements provided by relational distillation more precisely, we compute the Mean Absolute Error (MAE) between the similarity matrices across both intra- and cross-modal samples on ShapeNet to examine the effects of incorporating various relations. In \textbf{Tab.~\ref{tab:mae}}, it is evident that IR leads to a decrease in MAE between intra-modal similarities, while CR slightly reduces the divergence of cross-modal similarities. When DD is introduced, MRD achieves a better balance between multiple intra- and cross-modality relations. 

\begin{figure}[!t]
    \begin{minipage}[b]{0.49\textwidth}
        \centering
        \includegraphics[width=\textwidth]{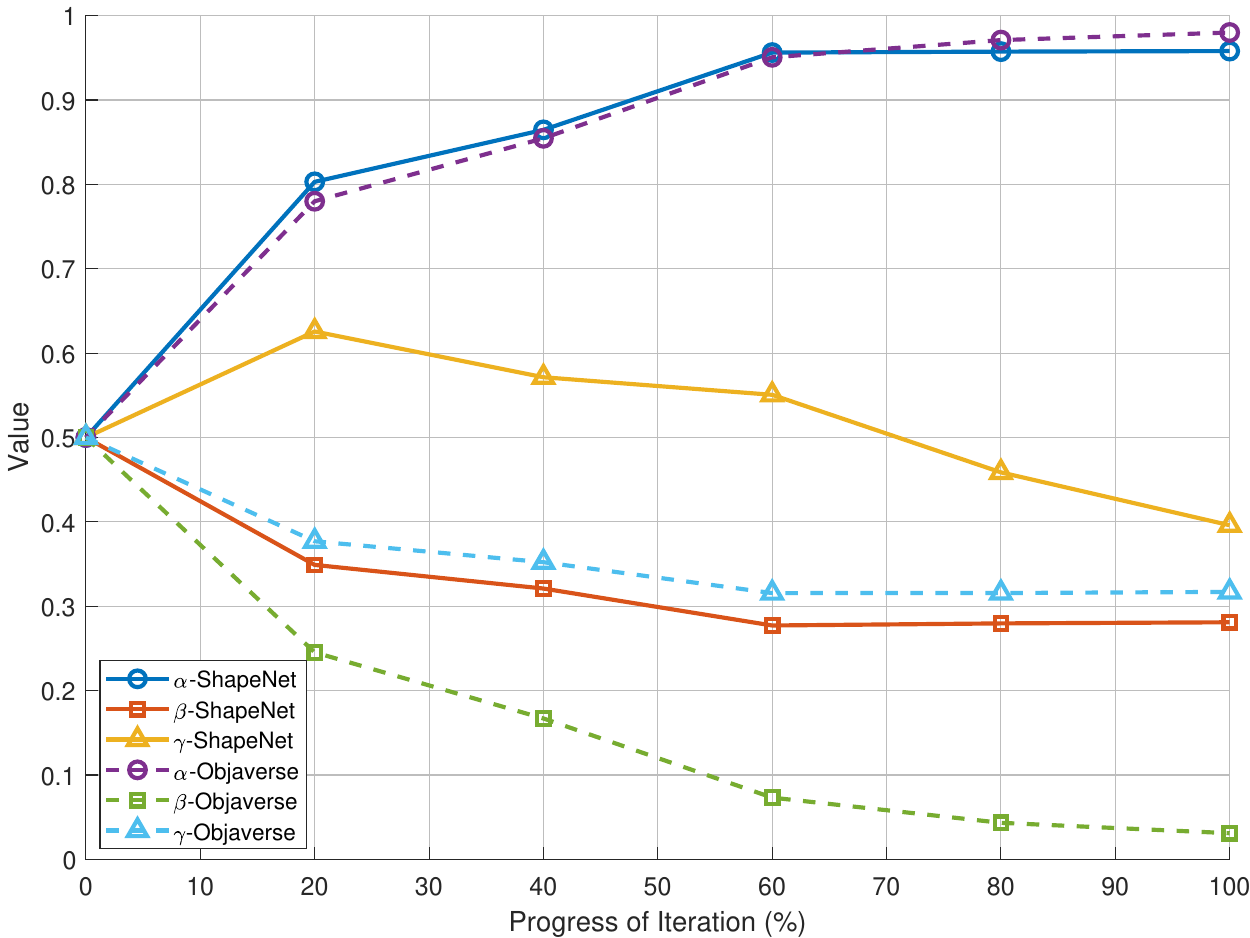}
        \caption{Visualization of the value changes of dynamic weights along with the progress of iterations on ShapeNet and Objaverse.}
        \label{fig:value}
    \end{minipage}
    \hfill
    \begin{minipage}[b]{0.49\textwidth}
        \centering
        \includegraphics[width=\textwidth]{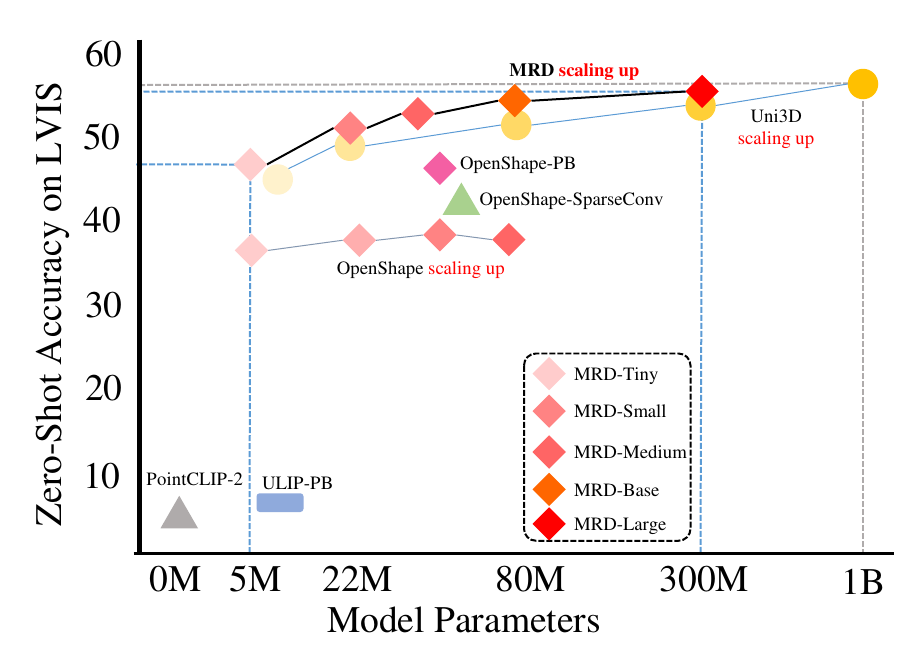}
        \caption{Comparison of model parameters and zero-shot accuracy (\%) on Objaverse, where MRD achieves the highest parameter efficiency.}
        \label{fig:scaling}
    \end{minipage}
\end{figure}

\textbf{Visualization of Dynamic Weights.} We visualize the value changes of the dynamic weights of MRD-M when trained on ShapeNet and Objaverse in Fig. \ref{fig:value}. As shown in Fig. \ref{fig:value}, the intra-modal relations tend to mimic the image-image relations, as they share common information such as textures and geometries between the 3D and image modalities. In contrast, the cross-modal relations lean towards synthesizing the text-image relations, likely due to the inherent differences between textual descriptions and visual appearnaces. Meanwhile, because of the difference of data distributions of ShapeNet and Objaverse, the dynamic weights converge to distinct values.

\textbf{Scaling up Model Size.}  We investigate the impact of scaling up the model size of PointBERT, similar to the way taken in Uni3D and OpenShape. The hyper-parameters of the model architecture and their corresponding accuracies are outlined in Tab. \ref{tab:scaleup}. Unlike the overfitting issue observed in OpenShape when scaling PointBERT parameters up to 72M, PointBERT trained with MRD successfully scales from 5M to 307M parameters. Additionally, as shown in Fig. \ref{fig:scaling}, despite that Uni3D utilizes a large-scale image dataset for pre-training, MRD still outperforms Uni3D with comparable model sizes, demonstrating its superiority. 

\section{Conclusion and Discussion}

We introduce MRD, an innovative approach to acquiring a unified multi-modal representation of 3D shapes. By distilling both intra-modal and cross-modal relations from the pre-aligned image and text modalities into the 3D modality, MRD facilitates the achievement of more discriminative representations. Our findings demonstrate notable improvements in the zero-shot shape classification and cross-modal retrieval tasks over previous research, underscoring the superior capability of MRD in 3D shape understanding.

Looking ahead, several avenues merit further investigation:  a) developing advanced methods for relationship representation and distillation for improved 3D and multi-modal features; b) enhancing the granularity of relationship characterization to extract richer semantic insights for more robust 3D representations; and c) improving conflict removal mechanisms to integrate diverse relational representations for deeper insight into representational relations across modalities.

\newpage

\section*{Acknowledgements}
This work is partly supported by the National Natural Science Foundation of China (No. 62176012 and 62022011), the Research Program of State Key Laboratory of Software Development Environment, and the Fundamental Research Funds for the Central Universities, with additional in-kind contributions from PICO ARCH.

%
%
\bibliographystyle{splncs04}
\bibliography{main}

\newpage
\appendix
\section{More Ablation Studies}
\textbf{Hyper-parameter.} We perform an ablation study to examine the impact of the tunable hyper-parameter $\lambda$ in Eq. (13). The results are depicted in Tab. \ref{tab:lambda}.

From Tab. \ref{tab:lambda}, we find that the optimal performance across all three benchmarks is achieved when $\lambda$ is set to 3. Consequently, we adopt $\lambda=3$ for all the experiments within this study.

\begin{table}[ht]
  \centering
  \caption{Ablation results (\%) on the Impact of the hyper-parameter $\lambda$.}
 
{
    \begin{tabular}{c|ccc|ccc|ccc}
    \toprule
        \multirow{2}[4]{*}{Setting}   & \multicolumn{3}{c|}{Objaverse-LVIS} & \multicolumn{3}{c|}{ModelNet40} & \multicolumn{3}{c}{ScanObjectNN} \\
\cmidrule{2-10}      & Top1  & Top3  & Top5  & Top1  & Top3  & Top5  & Top1  & Top3  & Top5 \\
    \midrule
    Base &  11.0   & 20.0      & 24.6   & 72.2  & 85.5      & 88.8  & 53.4  & 73.5      & 81.9 \\
    \midrule
    $\lambda=1$  &  \underline{11.8}      & \underline{21.2}      & \textbf{26.0}       & 72.9      & 87.8  & \underline{92.1}     & 55.5      & 75.2      & 83.5 \\
     $\lambda=2$ &   11.5   &  20.8    &   25.7    &   72.9    & \underline{88.0}   &  \textbf{92.8}     & \textbf{55.8}      & \underline{75.4}   & \underline{84.3}   \\
     $\lambda=3$ & \textbf{11.8} & \textbf{21.2} &25.8 &\textbf{74.2} &\textbf{88.3} & 90.9 &\underline{55.7} &\textbf{75.7} & 83.0 \\
      $\lambda=4$  & 11.5      & 20.9      & \underline{25.9}      & \underline{73.7}  & 87.6      & 91.5      & 53.7      & 74.9      & \textbf{84.4} \\
     $\lambda=5$  &   11.5     &   20.8    &  25.6       & 72.4 &   87.2   & 91.3 &  52.5     & 72.5   &  81.9    \\
    \toprule
    \end{tabular}}
  \label{tab:lambda}%
\end{table}%

\textbf{Downstream Finetuning.} We follow the same fine-tuning protocols as ULIP on standard 3D classification and present the results in \textbf{Tab.~\ref{tab:finetune}}. MRD largely surpasses the baseline and other counterparts in both the settings.

\begin{table}[t]
  \centering
  \caption{Results (\%) standard 3D classification on ScanObjNN. $*$ indicates the results obtained by re-implementation.}
    {\begin{tabular}{c|ccc|ccc}
    \toprule
     &\multicolumn{3}{c|}{PointBERT-22M}& \multicolumn{3}{c}{PointBERT-38M} \\
    \midrule
    Pretrained & - & ULIP & MRD & - & OpenShape & MRD \\
    \midrule
    Ins. Acc.& 83.1 &86.4 & 89.6 & 82.5* & 85.3* & 89.2 \\  
    \toprule
    \end{tabular}}
  \label{tab:finetune}
\end{table}

\begin{figure}[th]
\begin{center}

	\includegraphics[width = 1.0\linewidth]{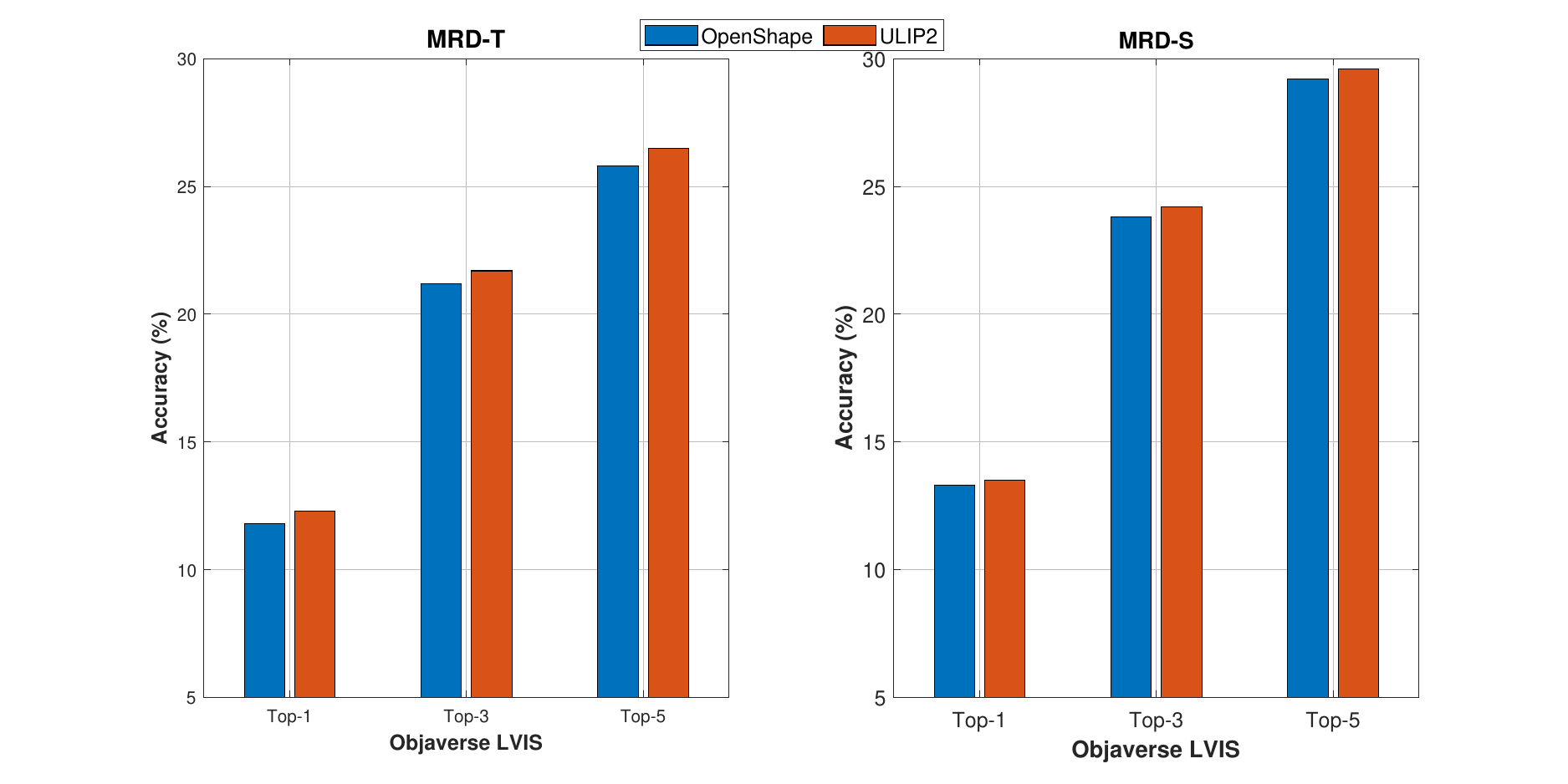}
   \caption{Comparison of zero-shot accuracy (\%) on Objaverse When trained with different data sources.}
   \label{fig:ulip}
   \end{center}
\end{figure}

\textbf{Data Source.} Since ULIP2 generates image-text pairs for 3D point clouds using different protocols, we train MRD-T and MRD-S on ShapeNet with the image-text pairs generated by ULIP2 to evaluate the data source generalization ability of MRD. The hyper-parameters are set as stated in Sec. 4.1, consistent with the previous ablation studies in the main text. The results are shown in Fig. \ref{fig:ulip}. As depicted, MRD-T and MRD-S achieve better performance when trained with the image-text data from ULIP2, demonstrating the generalization ability.

\section{Details in Cross-Modal Retrieval}

For quantitative assessment in Sec. 4.3, we use the Objaverse test set, denoted as $\{(p_i,l_i)\}_{i=1}^N$ as our evaluation set. The encoded 3D point clouds are employed as cross-modal queries to retrieve the corresponding detailed text descriptions $\{ t_i \}_{i=1}^N$ generated by Cap3D. A retrieval is deemed a success at the instance level if the retrieved item is the $i$-th item itself. Additionally, a more lenient criterion is applied to category-wise retrieval: if the retrieved item belongs to the same category $l_i$ as the target, it is also considered a successful case.

\begin{figure}[!tb]
\begin{center}
	\includegraphics[width = 0.8\linewidth]{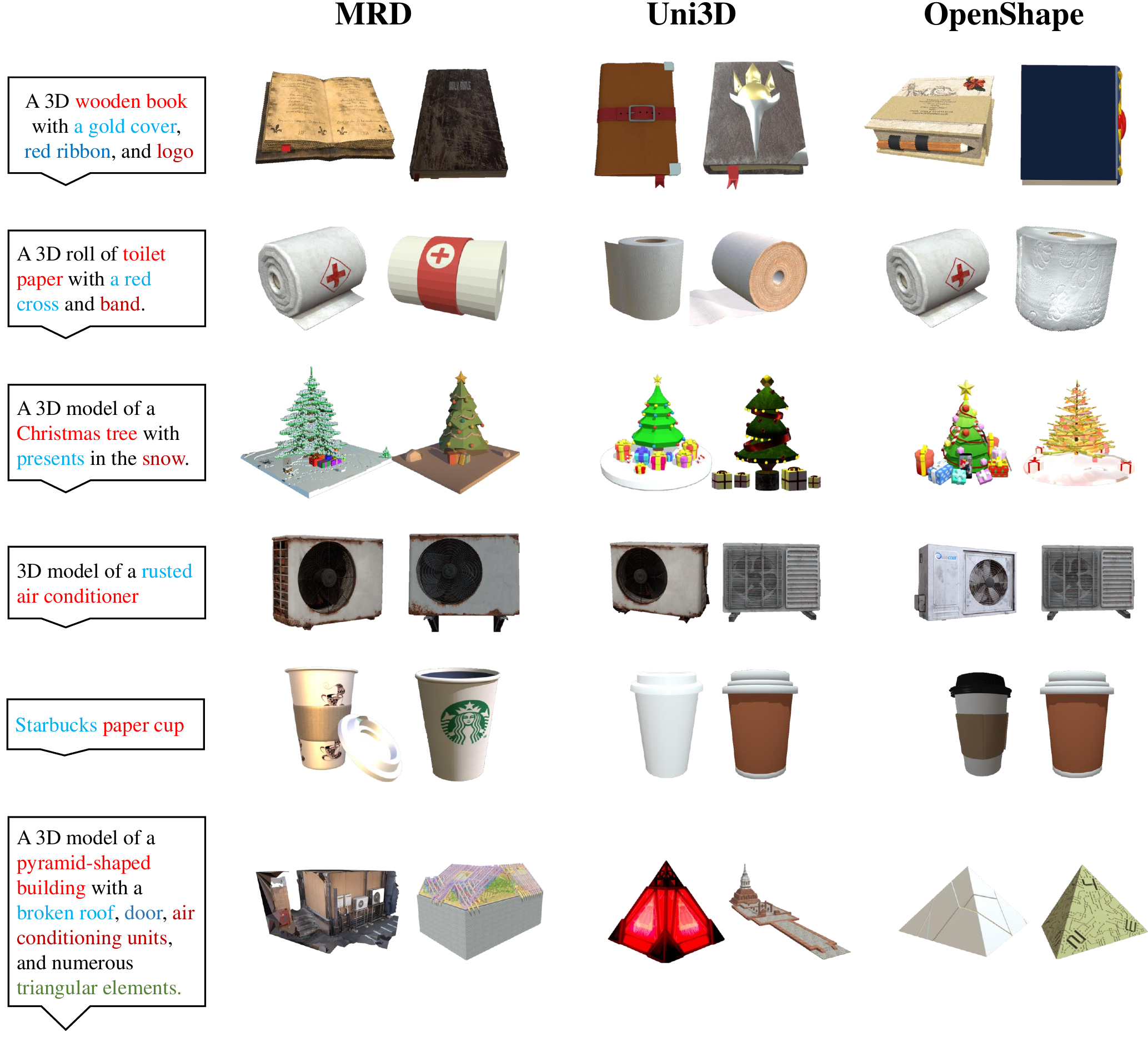}
 \end{center}
   \caption{Comparison of text-query 3D shape retrieval results. For every query text, we retrieve two 3D shapes that match most closely. Text descriptions feature words in different colors to highlight the diverse attributes of the retrieval targets. The results for MRD, Uni3D, and OpenShape are presented across columns 2 to 4.}
\label{fig:figt2p}
\end{figure}

For fair comparison, we utilize the released OpenShape-PointBert and Uni3D-B models as well as MRD-B, which are all on a comparable scale in terms of parameters. The retrieval accuracy is reported in Tab.2 in the main text. Additionally, more qualitative comparison is visualized in Fig. \ref{fig:figt2p}, wherer we can find that MRD achieves the superior performance compared to the counterparts.

\begin{figure}[!tb]
\begin{center}
	\includegraphics[width = 1.0\linewidth]{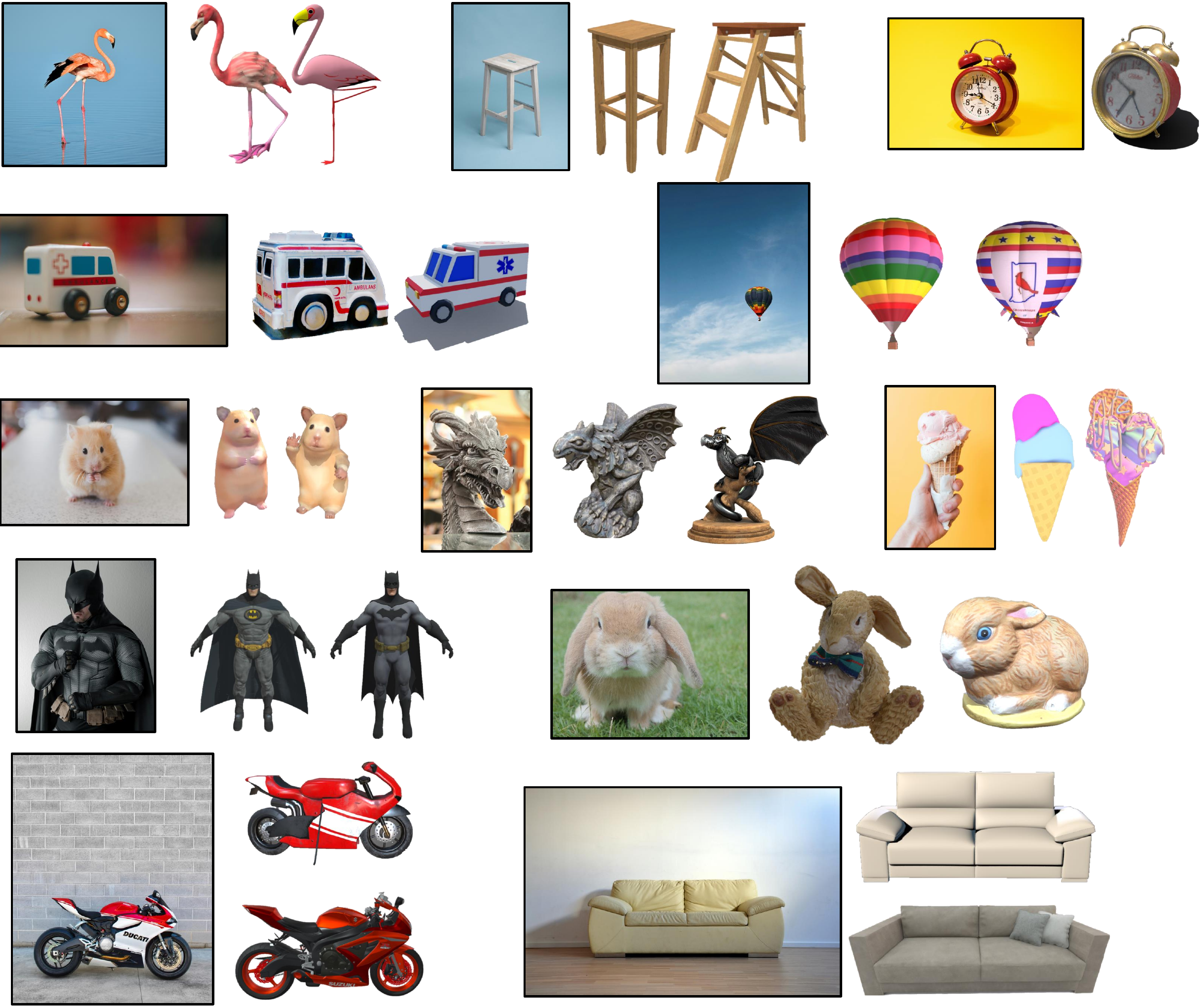}
 \end{center}
   \caption{Visualization of more image-query 3D shape retrieval results. Input images are from unsplash.com.}
\label{fig:figi2p}
\end{figure}

We further visualize additional image-query 3D shape results generated by MRD in Fig. \ref{fig:figi2p}. 

\end{document}